\documentclass[10pt,twocolumn,letterpaper]{article}

\usepackage{iccv}
\usepackage{times}
\usepackage{epsfig}
\usepackage{graphicx}
\usepackage{amsmath}
\usepackage{amssymb}

\usepackage{graphicx}
\usepackage{amsmath}
\usepackage{amssymb}
\usepackage{booktabs}
\usepackage{multirow}
\usepackage{gensymb}
\usepackage[shortcuts]{extdash}
\usepackage{url}
\usepackage[table]{xcolor}
\usepackage[breaklinks=true,bookmarks=false]{hyperref}
\iccvfinalcopy 

\def\iccvPaperID{6133} 

\ificcvfinal\fi

\begin{document}

\title{RegFormer: An Efficient Projection-Aware Transformer Network for Large-Scale Point Cloud Registration}

\author{Jiuming~Liu\textsuperscript{\rm 1}, Guangming~Wang\textsuperscript{\rm 1}, Zhe~Liu\textsuperscript{\rm 2}$^{*}$, Chaokang~Jiang\textsuperscript{\rm 3}, Marc~Pollefeys\textsuperscript{\rm 4,5}, Hesheng~Wang\textsuperscript{\rm 1}\thanks{ Corresponding Authors. 
		\indent~ The first two authors contributed
		equally.
	}\\
	{\textsuperscript{\rm 1}Department of Automation, Key Laboratory of System Control}\\
{and Information Processing of Ministry of Education, Shanghai Jiao Tong University}\\
    {\textsuperscript{\rm 2} MoE Key Lab of Artificial Intelligence, AI Institute, Shanghai Jiao Tong University}\\
     {\textsuperscript{\rm 3}
		China University of Mining and Technology} 
    {\textsuperscript{\rm 4}
		ETH Zürich}  
     {\textsuperscript{\rm 5}
		Microsoft}  ~~
 \\
	\small{\texttt{\{liujiuming,wangguangming,liuzhesjtu,wanghesheng\}@sjtu.edu.cn}} \\
	\small{\texttt{ts20060079a31@cumt.edu.cn}}\qquad \small{\texttt{marc.pollefeys@inf.ethz.ch}}
}

\maketitle
\ificcvfinal\thispagestyle{empty}\fi

\begin{abstract}
   Although point cloud registration has achieved remarkable advances in object-level and indoor scenes, large-scale registration methods are rarely explored. Challenges mainly arise from the huge point number, complex distribution, and outliers of outdoor LiDAR scans. In addition, most existing registration works generally adopt a two-stage paradigm: They first find correspondences by extracting discriminative local features and then leverage estimators (eg. RANSAC) to filter outliers, which are highly dependent on well-designed descriptors and post-processing choices. To address these problems, we propose an end-to-end transformer network (RegFormer) for large-scale point cloud alignment without any further post-processing. Specifically, a projection-aware hierarchical transformer is proposed to capture long-range dependencies and filter outliers by extracting point features globally. Our transformer has linear complexity, which guarantees high efficiency even for large-scale scenes. Furthermore, to effectively reduce mismatches, a bijective association transformer is designed for regressing the initial transformation. Extensive experiments on KITTI and NuScenes datasets demonstrate that our RegFormer achieves competitive performance in terms of both accuracy and efficiency. Codes are available at \url{https://github.com/IRMVLab/RegFormer}.
\end{abstract}


\begin{figure}
  \centering
  \includegraphics[width=1.00\linewidth]{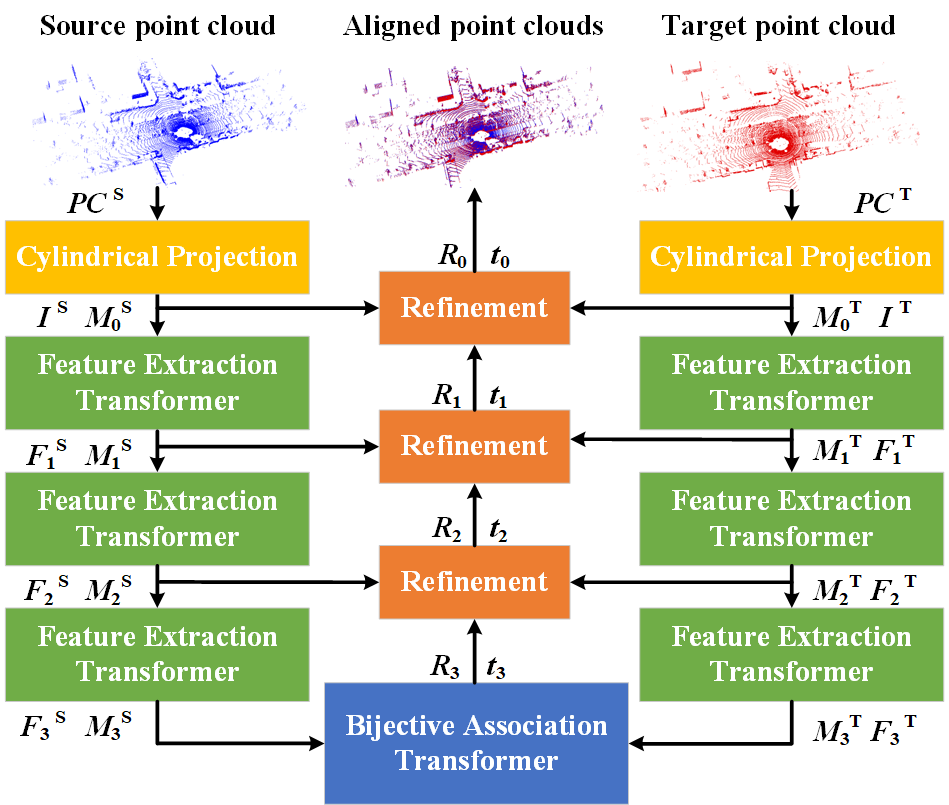}
  \caption{Overview architecture of RegFormer. The whole feature extraction and frame association sections are transformer-based. We project point cloud onto a 2D surface and feed its patches into transformer. A projection mask $M^T$($M^S$) is also proposed, which equips our transformer with the awareness of invalid positions.}
  \label{fig:pipeline}
  \vspace{-12pt}
\end{figure}

\section{Introduction}
\label{sec:intro}
Point cloud registration is a fundamental problem in 3D computer vision, which aims to estimate the rigid transformation between point cloud frames. It is widely applied in moblie robotics \cite{liu2023translo,wang2022residual}, autonomous driving \cite{zheng2022global,wang2022efficient}, etc. 


Although learning-based methods show great potential in object-level or indoor registration tasks \cite{yew20183dfeat,choy2019fully,ao2021spinnet,huang2021predator}, large-scale point cloud registration is less studied. Challenges are mainly three-fold: 1) Outdoor LiDAR scans may consist of hundreds of thousands of unstructured points, which are intrinsically sparse, irregular, and have a large spatial range. It is non-trivial to efficiently process all raw points in one inference \cite{yew2022regtr}. 2) Outliers from dynamic objects and occlusion would degrade the registration accuracy as they introduce uncertain motions and inconsistency. 3) There are numerous mismatches when directly leveraging distance-based nearest neighbor matching methods (eg. $k$NN) to distant point cloud pairs \cite{lu2021hregnet}.

For the first challenge, previous registration works mostly voxelize input points \cite{bai2020d3feat,lu2021hregnet}, and then establish putative correspondences by selecting keypoints and learning distinctive local descriptors \cite{yew20183dfeat,choy2019fully,ao2021spinnet}. However, quantization errors are inevitable in the voxelization \cite{huang20223qnet}. Also, different selected keypoints may influence registration accuracy and downsampling challenges the repeatability \cite{yu2021cofinet}. In this paper, instead of searching keypoints, we directly process all LiDAR points by projecting them onto a cylindrical surface for the structured organization. Projected image-like structure facilitates the window partition in transformer, realizing linear computational costs. This enables our network to process almost 120000 points with high efficiency. To take advantage of 3D geometric features, each projected position is filled with raw point coordinates, inspired by \cite{wang2022efficient}. Another concern is that projected pseudo images are full of invalid positions due to the original sparsity of point clouds. We handle this by designing a projection mask.  

For the second challenge, the commonly used method is applying the robust estimator (RANSAC) \cite{fischler1981random,bai2020d3feat,ao2021spinnet} to filter outliers. However, RANSAC suffers from slow convergence \cite{qin2022geometric} and is highly dependent on post-processing choices \cite{yew2022regtr}. From a different view, we observe that global modeling capability is rather helpful to localize occluded objects and recognize dynamics as they introduce inconsistent global motion. Therefore, we propose a projection-aware transformer to extract point features globally. Notably, some recent works \cite{qin2022geometric,yew2022regtr} also try to design RANSAC-free registration networks. However, the combination of CNN and transformer in their feature extraction modules deteriorates the efficiency. The closest approach to ours is REGTR \cite{yew2022regtr}, which directly predicts clean correspondences with transformer. Nonetheless, the quadratic complexity limits its ability for large-scale application.


In addition, a Bijective Association Transformer (BAT) is designed to tackle the third challenge. HRegNet \cite{lu2021hregnet} already has awareness that nearest-neighbor matching can lead to considerable mismatches due to possible errors in descriptors. However, their $k$NN cluster is still distance-based, which can not generalize well to low-overlap inputs. To address this problem, two effective components are designed in BAT for reducing mismatches. The cross-attention mechanism is utilized first for preliminary location information exchange. Intuitively, features of deeper layers are coarse but reliable as they gather more information with larger receptive fields. Thus, each point is correlated with all points (instead of selecting $k$ points) in the other frame to gain reliable motion embeddings on the coarsest layer (all-to-all). The precise transformation will then be recovered by the iterative refinement on shallow layers.

\textbf{Overall, our contributions are as follows:}
    \vspace{-0.3cm}
\begin{itemize}
	\item We propose a fully end-to-end network for large-scale point cloud registration. It does not need any keypoint matching or post-processing, which is both keypoint-free and RANSAC-free. Our efficient model can process hundreds of thousands of points in real time.
	\vspace{-0.3cm}
	\item The global modeling capability of our RegFormer can filter outliers effectively. Furthermore, a Bijective Association Transformer (BAT) is designed to reduce mismatches by combining cross-attention with an all-to-all point correlation strategy on the coarsest layer. 
	\vspace{-0.7cm}
	\item Experiment results on KITTI \cite{geiger2012we,geiger2013vision} and NuScenes \cite{caesar2020nuscenes} datasets indicate that our RegFormer achieves state-of-the-art performance with 99.8\% and 99.9\% successful registration recall respectively.
	
\end{itemize}

\section{Related Work}
\textbf{Deep point cloud registration.} Existing deep point cloud registration networks can be divided into two categories according to whether they extract explicit correspondences. The first class attempts to establish point correspondences through keypoint detection \cite{lu2019deepvcp,fu2021robust,lu2021hregnet} and learning powerful discriminative descriptors \cite{yew20183dfeat,choy2019fully,bai2020d3feat,ao2021spinnet,deng2018ppfnet}. As a pioneering work, 3DMatch \cite{zeng20173dmatch} extracts local volumetric patch features by a Siamese network. PPFNet \cite{deng2018ppfnet} and its unsupervised version PPF-FoldNet \cite{deng2018ppf} extract global context-aware descriptors using PointNet \cite{qi2017pointnet}. To enlarge the receptive field, FCGF \cite{choy2019fully} computes dense descriptors of whole input point clouds in a forward pass. Recent correspondence-based networks \cite{choy2020deep, yang2020teaser,bai2021pointdsc,chen2022deterministic,chen2022sc2} commonly use it to generate putative correspondences.

The second class directly estimates transformation in an end-to-end manner \cite{aoki2019pointnetlk,wang2019deep,huang2020feature,yew2020rpm}. Point clouds are aligned with learned soft correspondences or without explicit correspondences. Among these, PointNetLK \cite{aoki2019pointnetlk} is a landmark that extracts global descriptors and estimates transformation with Lucas-Kanade algorithm \cite{lucas1981iterative}. FMR \cite{huang2020feature} enforces the optimization of registration by minimizing feature-metric projection errors. However, these direct registration methods can not generalize well to large-scale scenes \cite{huang2021predator}. Our method falls into this category and is specially designed for large-scale registration.

\textbf{Large-scale point cloud registration.} Large-scale registration is less studied in previous works. DeepVCP \cite{lu2019deepvcp} incorporates both the local similarity and global geometric constraints in an end-to-end manner. Although it is evaluated on outdoor benchmarks, its keypoint matching is still constrained in local space. With a larger keypoint detection range, HRegNet \cite{lu2021hregnet} introduces bilateral and neighborhood consensus into keypoint features and achieves state-of-the-art. Different from previous works \cite{lu2019deepvcp,lu2021hregnet,wiesmann2022dcpcr} that mostly focus on local descriptors, we address the issue from a more global perspective thanks to the long-range dependencies capturing ability of transformer. Motivated by recent scene flow works \cite{wang2022sfgan,wang2021unsupervised}, our RegFormer has no need for searching keypoint and explicit point-to-point correspondences, which learns implicit cross-frame motion and directly outputs pose in a single pass.

\textbf{Transformer in registration tasks.} Most existing works \cite{wang2019deep,fu2021robust,yew2022regtr,yu2021cofinet} only treat transformer as a frame association module. Among these, DCP \cite{wang2019deep} first utilizes a vanilla transformer to correlate downsampled features. RGM \cite{fu2021robust} proposes a framework based on deep graph matching, where transformer is employed to dynamically learn the soft edges of nodes. REGTR \cite{yew2022regtr} outputs overlap scores and the location information through cross-attention, but it can not handle large-scale scenes. Our RegFormer achieves linear complexity by revisiting attention within non-overlapping windows. Transformer is designed for not only frame association but also feature extraction. To the best of our knowledge, our RegFormer is the first pure transformer-based registration network.

\begin{figure}
  \centering
  \includegraphics[width=1.00\linewidth]{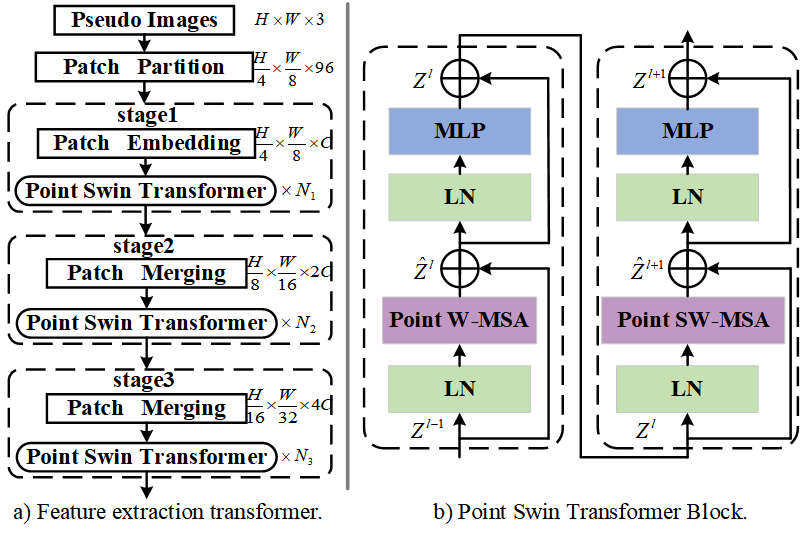}
  \vspace{-14pt}
  \caption{The feature extraction module consists of three cascaded stages as constructed in a). b) indicates Point Swin Transformer block in stage $l$, which computes attention within windows (Point W-MSA), and then gathers contextual information by the spatial shift (Point SW-MSA).}
  \label{fig:feature extraction}
  \vspace{-12pt}
\end{figure}

\section{RegFormer}
\subsection{Overall Architecture}
\label{overall}
 The overall architecture of our proposed RegFormer is illustrated in Fig.~\ref{fig:pipeline}. Given two point cloud frames: source point cloud $PC^{S}\in\mathbb{R}^{N\times3}$ and target point cloud $PC^{T}\in\mathbb{R}^{M\times3}$, the objective of registration is to align them via an estimated transformation. To orderly organize raw irregular points, we first project point clouds as pseudo images $I^{S}$ and $I^{T}$ in Section~\ref{sec:projection}, and then feed them together with corresponding masks $M_{0}^{S}$ and $M_{0}^{T}$ into the hierarchical feature extraction module as in Fig.~\ref{fig:feature extraction} a). Following prior works \cite{liu2021swin,liu2022video}, we treat each patch of size $4\times 8\times 3$ as a token, and then a feature embedding layer is utilized to project these patches to an arbitrary dimension denoted by $C$. The patch merging layer of each stage concatenates $2\times 2$ neighbor patches. Then, concatenated features are reduced to half channels, and then fed into a projection-aware transformer in Section~\ref{pst}. For associating point clouds and reducing mismatches, a Bijective Association Transformer (BAT) in Section~\ref{bat} is employed to generate initial motion embeddings. Finally, the quaternion $q_{3}\in\mathbb{R}^{4}$ and translation vector $t_{3}\in\mathbb{R}^{3}$ are estimated from motion embeddings, and then refined iteratively.

\subsection{Cylindrical Projection}
\label{sec:projection}
\begin{figure}
 \centering
 \includegraphics[width=1.0\linewidth]{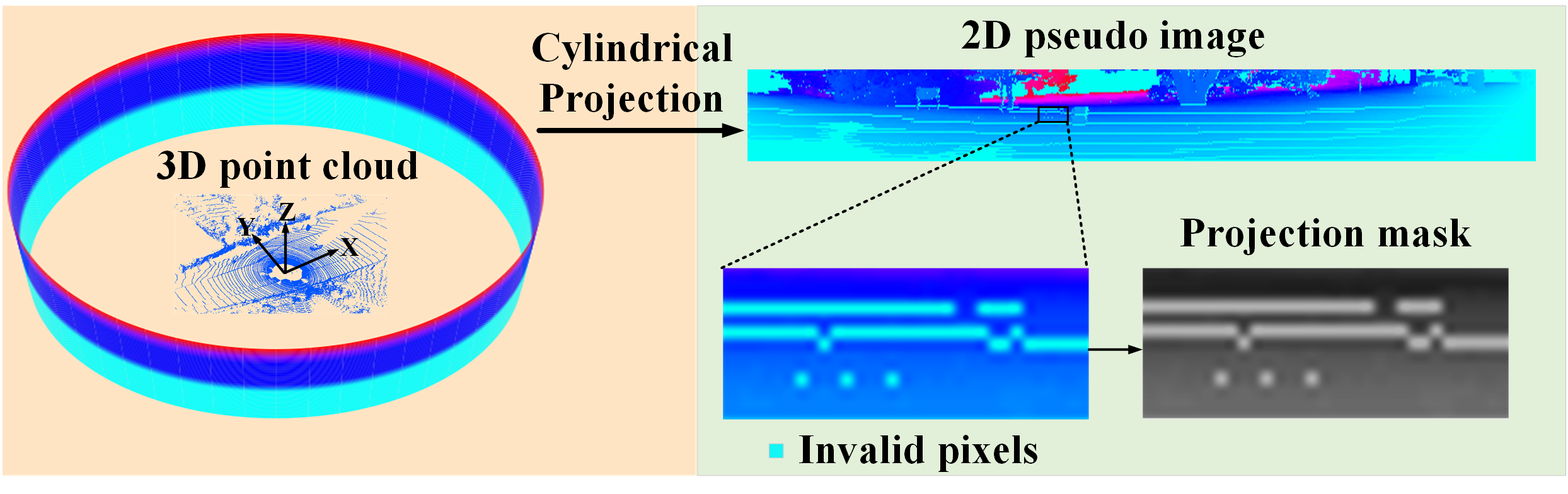}
 \vspace{-14pt}
 \caption{Cylindrical projection. We project 3D point clouds onto a 2D surface and fill each pixel with its raw $x,y,z$ coordinates. A projection mask is also proposed to remove invalid positions.}
 \vspace{-8pt}
 \label{fig:pro}
\end{figure}

According to the original proximity relationship of raw points, point clouds are projected onto a cylindrical surface, following the line scanning characteristic of the LiDAR sensor. Each point has a corresponding 2D pixel position on projected pseudo images as:
\vspace{-0.2cm}
\begin{equation}
    \label{eq:projection1}
     u = arctan 2(y/x)/\Delta\theta,
\end{equation}
\vspace{-0.4cm}
\begin{equation}
\label{eq:projection2}
     v = arcsin (z/\sqrt{x^2 + y^2 +z^2})/\Delta\phi,
\end{equation}

where $x,y,z$ represent the raw 3D coordinates of point cloud and $u,v$ are corresponding 2D pixel positions. $\Delta\theta,\Delta\phi$ are horizontal and vertical resolutions of the LiDAR sensor. To make the best use of geometric information of raw 3D points, we fill each pixel position with its raw $x,y,z$ coordinates. Pseudo images of size $H\times W\times 3$ in Fig.~\ref{fig:pro} will be input to the feature extraction transformer. 


\subsection{Point Swin Transformer} 
\label{pst}
Compared with images, the scale of outdoor LiDAR points is surprisingly larger, and thus they require a much larger number of tokens for representation. Vanilla transformer with quadratic complexity is not suitable, as it will lead to huge computation costs. Inspired by Swin Transformer \cite{liu2021swin}, we introduce window attention into 3D point transformer for linear complexity. Thanks to the global understanding ability of transformer, our network can effectively learn to identify dynamic motions and the location of occluded objects in the other frame. To simplify the formulation, we only expand the description for the source point cloud in this section, and the same goes for the target one.

\begin{figure*}
 \centering
 \includegraphics[width=1.0\linewidth]{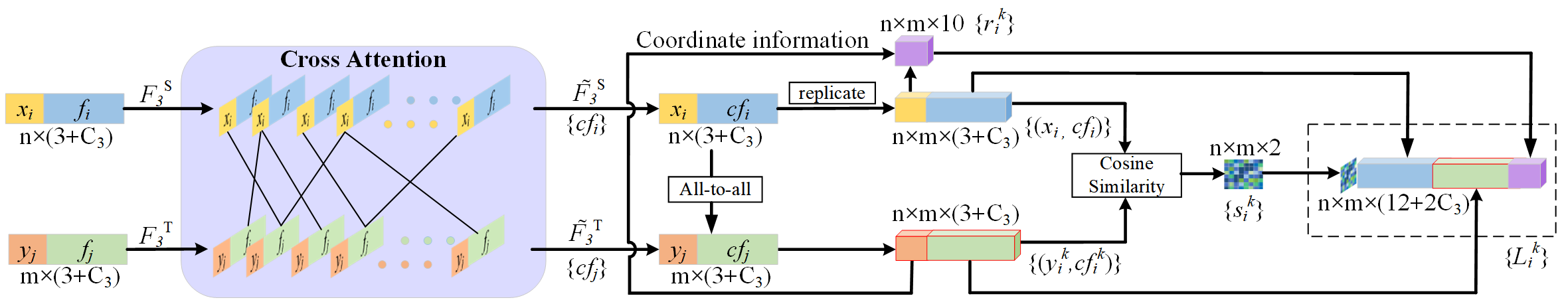}
 \vspace{-14pt}
 \caption{Bijective Association Transformer. The cross-attention mechanism is leveraged for preliminary information exchange between two frames. Then, geometric characteristics of conditioned features ${\tilde{F}}^{S}_{3}, {\tilde{F}}^{T}_{3}$ are fully considered to generate the initial motion embeddings.}
 \vspace{-10pt}
 \label{fig:bat}
\end{figure*}

\textbf{Projection masks.} It is non-trivial to extend the 2D window-based attention mechanism to the pseudo images generated from 3D points. Point cloud, especially in outdoor scenes, is extremely sparse. Thus, projected pseudo images are filled with invalid blank pixels. The registration accuracy will be affected if they are fed identically into attention. Also, the attention calculation itself of these pixels is meaningless as they have no corresponding raw points. Inspired by \cite{cheng2021per}, a projection-aware mask $M_{l}^{S}$ is proposed here, which represents whether each pixel is invalid in Fig.~\ref{fig:pro}:
\begin{equation}
\vspace{-0.3cm}
M_{l}^{S}=
\begin{cases}
-\infty,& \text{ $ x = 0,y = 0,z = 0 $ }, \\
0,& \text{ $ otherwise $ },
\end{cases}
\end{equation}
where $x,y,z$ are point coordinates filled into pseudo images. The projection mask $M_{l}^{S}$ of size $H\times W\times 1$ is pixel-corresponding to the projected pseudo image and together downsampled in each stage as Fig.~\ref{fig:pipeline}. $l$ denotes the stage number. We assign zero to valid pixels and a big negative number to invalid ones where attention should not be calculated. In this way, invalid pixels would then be filtered through the softmax operation afterward in attention blocks.


\textbf{Point W-MSA and Point SW-MSA.} For stage $l$, point feature $Z^{l-1}$ ($H_{l}\times W_{l}\times C_{l}$) and its corresponding mask $M_{l}^{S}$ ($H_{l}\times W_{l}\times 1$) are fed into Point Window-based Multi-Head Self Attention (Point W-MSA) as: 
\vspace{-0.2cm}
\begin{equation}
    \label{eq:WMSA}
    W\-/MSA(Z^{l-1}) = (Head_1 \oplus \cdots \oplus Head_H)W^O, 
\end{equation}
\vspace{-24pt}
\begin{align}
      \label{eq:head}
               Head_h 
              &= Attention({Q^h}, {K^h}, {V^h}) \notag\\ 
              &\vspace{-10pt}= softmax(\frac{{Q^h}{K^h}}{\sqrt{d_{head}}}+ M_{l}^{S}+ Bias)V^h ,
      \vspace{-5pt}
\end{align}
where $Head_h$ represents the output of $h\-/th$ head. $M_{l}^{S}$ is the projection mask. $Bias$ is the relative position encoding \cite{raffel2020exploring}. $Q^h=Z^{l-1}{W_h^Q}$, $K^h=Z^{l-1}{W_h^K}$, ${V^h}=Z^{l-1}{W_h^V}$, in which $W_h^Q\in{\mathbb{R}^{ C_{l} \times C_{head}}}$, $W_h^K\in{\mathbb{R}^{ C_{l} \times C_{head}}}$, $W_h^V\in{\mathbb{R}^{ C_{l} \times C_{head}}}$, $W^O\in{\mathbb{R}^{H C_{head} \times C_{l}}}$ are learned projections. The above process is repeated again in the following Point Shift Window-based Multi-head Self Attention (Point SW-MSA). The only difference is that features are first spatially shifted \cite{liu2021swin} in Point SW-MSA for increasing information interaction among windows.

\textbf{Point Swin Transformer blocks.} Overall, one complete transformer stage in Fig.~\ref{fig:feature extraction} b) can be described as:\vspace{-0.2cm}
\begin{align}
    \label{transformer}
    &{\hat{Z}}^l = PW\-/MSA(LN(Z^{l-1})) + Z^{l-1} \notag\\
    &{Z^l} = MLP(LN({\hat{Z}}^l)) + {\hat{Z}}^l \notag\\
    &{\hat{Z}}^{l+1} = PSW\-/MSA(LN(Z^{l})) + Z^{l} \notag\\
    &{Z^{l+1}} = MLP(LN({\hat{Z}}^{l+1})) + {\hat{Z}}^{l+1},
\end{align}
where P(S)W-MSA represents Point (Shift) Window Multi-head Self Attention. ${Z}^{l+1}$ is the output feature of stage $l$.

\subsection{Bijective Association Transformer}
\label{bat}
After global features are hierarchically extracted by our Point Swin Transformer, the key issue is how to match source and target point clouds through their downsampled features. The most common method is to search for nearest neighbors (NN). However, this distance-dependent strategy is ineffective enough for large-scale registration, as two corresponding points may be too far away, which leads to numerous mismatches \cite{lu2021hregnet}. To solve this problem, we propose a Bijective Association Transformer block (BAT) in Fig.~\ref{fig:bat}, which first learns rough but generally correct location information with cross-attention. Then, an all-to-all point gathering strategy guarantees reliable location output and further reduces mismatches on the coarsest layer.

\textbf{Rough association.} As depicted in Fig.~\ref{fig:bat}, downsampled source and target point features of stage 3 are first fed into a cross-attention layer with linear complexity, roughly associating with each other. Cross attention can introduce certain similarities of two point cloud frames by calculating attention weights and updating features with the awareness of point location in the other frame \cite{yew2022regtr}. Concretely, source and target point coordinates and their features are first resized as ${X^S_3}\in\mathbb{R}^{n\times3}$, ${Y^T_3}\in\mathbb{R}^{m\times3}$ and ${F^S_3}\in\mathbb{R}^{n\times C_{3}}$, ${F^T_3}\in\mathbb{R}^{m\times C_{3}}$, which are inputs of the cross-attention block. The output conditioned features ${\tilde{F}}^{S}_3$ for source point cloud can be written as:\vspace{-0.1cm}
\begin{equation}
    \label{eq:cross}
   {\tilde{F}}^{S}_{3}
   = Attention({F^{S}_{3}{W^Q}}, {F^{T}_{3}{W^K}}, {F^{T}_{3}{W^V}}),
\end{equation}
where $W^Q,W^K, W^V$ are respective projected functions. $Attention$ is similar to Section~\ref{pst}. When the source point cloud serves as $query$, the target point cloud would be projected as $key$ and $value$, and vice versa.

\textbf{All-to-all point gathering.} The coarsest layer obviously gathers more information and a larger receptive field, which is reliable to match two frames. Thus, on the bottom layer of our RegFormer, each point in ${\tilde{F}}^{S}_{3}$  is associated with all points in ${\tilde{F}}^{T}_{3}$, rather than select $k$ nearest neighbor points ($k$NN), to generate reliable motion embeddings.  Specifically, each point in ${PC^S}=\{(x_i,f_i)|{x_i}\in{X^S},{cf_i}\in{\tilde{F}}^{S}_{3}, i=1,\cdots,n\}$ correlates with all $m$ points in ${PC^T}=\{(y_j,f_j)|{y_j}\in{Y^T},{cf_j}\in{\tilde{F}}^{T}_{3}, j=1,\cdots,m\}$, forming an association cluster $\{(y_{i}^{k},cf^{k}_{i})|k=1,\cdots,m\}$. Then, the relative 3D Euclidean space information $\{r^{k}_{i}\}$ is calculated as:\vspace{-0.2cm}
\begin{equation}
    \label{eq:dis}
    r^{k}_{i} = {x_i} \oplus {y^{k}_{i}} \oplus (x_i - y^{k}_{i}) \oplus \Vert x_i - y^{k}_{i} \Vert_{2},
\end{equation}
where $\left\|  \cdot  \right\|_2$ indicates the $L_2$ Norm. 

The cosine similarity of grouped features is also introduced as:\vspace{-0.2cm}
\begin{equation}
    \label{eq:feat}
    s^{k}_{i} = \frac{<cf_i,cf_{i}^{k}>}{\Vert cf_i\Vert_{2} \Vert cf^{k}_{i}\Vert_{2}},
\end{equation}
where $< ,>$ denotes the inner product. This step will output a $n\times m\times 2$ similarity feature $s^{k}_{i}$, where the neighbor similarity in \cite{lu2021hregnet} is also considered here.

Then, we concatenate the above space and similarity embeddings and utilize a 3-layer shared MLP on them:\vspace{-0.2cm}
\begin{equation}
    \label{eq:cat}
    L^{k}_{i} = MLP(f_{i}\oplus cf^{k}_{i}\oplus r^{k}_{i}\oplus s^{k}_{i}).
\end{equation}

Finally, the initial flow embedding can be represented by the attentive encoding of concatenated features as:\vspace{-0.2cm}
\begin{equation}
    \label{eq:flow1}
    fe_{i} = \sum\limits_{k =1}^{m} {L^{k}_{i}} \odot{\mathop{softmax}\limits_{k =1,\cdots,m}{(L^{k}_{i})}},\vspace{-0.2cm}
\end{equation}
where a max-pooling layer and a softmax function are leveraged to predict attention weights for each point $x_i$. And the output motion embedding is a weighted sum of $ L^{k}_{i}$.
\subsection{Estimation of the Rigid Transformation} 
The transformation is estimated from initial motion embeddings $FE =\{fe_{i},i=1,\cdots,n\}$ together with downsampled source point features $F_3^S$ in layer 3 as:\vspace{-0.2cm}
\begin{equation}
    \label{eq:pose}
    W = softmax(MLP(FE \oplus {F_3^S})),
\end{equation}
where $W=\{w_i|w_i\in \mathbb{R} ^{C_{3}}\}$ are attention weights. Then, the quaternion $q_{3}\in\mathbb{R}^{4}$ and translation vector $t_{3}\in\mathbb{R}^{3}$ can be generated separately from weighting and sum operations followed by a fully connected layer:
\vspace{-0.2cm}
\begin{equation}
    \label{eq:pose1}
   q_3 = \frac{FC_1(\sum\limits_{i=1}^{n} fe_{i}\odot w_i)}{|FC_1(\sum\limits_{i=1}^{n} fe_{i}\odot w_i)|},
\end{equation}
\vspace{-0.2cm}
\begin{equation}
   \label{eq:pose2}
   t_3 = FC_2(\sum\limits_{i=1}^{n} fe_{i}\odot w_i),
\end{equation}
where $FC_1$ and $FC_2$ denote two fully connected layers.

Nonetheless, the initially estimated transformation is not precise enough due to the sparsity of the coarsest layer. Thus, we iteratively refine it on upper layers with PWC structure \cite{sun2018pwc, wang2021pwclo} to generate residual transformation $\Delta q^l$ and $\Delta t^l$. Refinement in the $l\-/th$ layer can be indicated as:
\vspace{-0.2cm}
\begin{equation}
   \label{eq:refine}
   q^l = \Delta q^l q^{l+1},
\end{equation}
\vspace{-0.5cm}
\begin{equation}
   \label{eq:refine1}
   [0,t^l] = \Delta q^l [0,t^{l+1}] ({\Delta q^l})^{-1}+[0,\Delta t^l].
\end{equation}

\subsection{Loss Function}
Our network outputs transformation parameters from four layers and adopts a multi-scale supervised approach:
${L}=\alpha^l  \mathcal{L}^l$. $\alpha^l$ indicates weights of layer $l$. $\mathcal{L}^l$ denotes the loss function of the $l\-/th$ layer, which is calculated as:
\vspace{-0.2cm}
\begin{equation}
   \mathcal{L}^l= \mathcal{L}_{trans}^l exp(-k_t) + k_{t} + \mathcal{L}_{rot}^l exp(-k_r) + k_{r},
\end{equation}
where $k_{t}$ and $k_{r}$ are two learnable parameters, which can uniform the difference in the unit and scale between quaternion and translation vectors \cite{li2019net}. $\mathcal{L}_{trans}^l$ and $\mathcal{L}_{rot}^l$ can be calculated as:
\vspace{-0.2cm}
\begin{equation}
   \label{eq:loss}
   \mathcal{L}_{trans}^l= \Vert t^l-\hat{t}^l\Vert,
\end{equation}
\vspace{-0.5cm}
\begin{equation}
   \label{eq:loss1}
   \mathcal{L}_{rot}^l= \Vert \frac{q^l}{\Vert q^l\Vert}-\hat{q}^l\Vert_2,
\end{equation} 
where $\left\|  \cdot  \right\|$ indicates the $L_1$ Norm. $q^l, t^l$ and $\hat{q}^l,\hat{t}^l$ are estimated and ground truth transformations respectively.
\section{Experiment}
We evaluate our RegFormer on two large-scale point cloud datasets, namely KITTI \cite{geiger2012we,geiger2013vision} and NuScenes \cite{caesar2020nuscenes}. Moreover, ablation studies are conducted for each designed component of our network to demonstrate their effectiveness. Extensive experiments demonstrate that our methods can achieve state-of-the-art registration accuracy and also guarantee high efficiency.

\subsection{Experiment Settings}
\textbf{Implement Details.} In the data processing, we directly input all LiDAR points without downsampling. Projected pseudo images are set in line with the range of LiDAR sensor as 64 ($H$)$\times$1792 ($W$) for KITTI and 32 ($H$)$\times$1792 ($W$) for NuScenes. Window size and shift size are set as 4 and 2 separately. Experiments are conducted on a single NVIDIA Titan RTX GPU with PyTorch 1.10.1. The Adam optimizer is adopted with $\beta_{1}$ = 0.9, $\beta_{2}$ = 0.999. The initial learning rate is 0.001 and exponentially decays every 200000 steps until 0.00001. The batch size is set as 8. The hyperparameter $\alpha^l$ in the loss function is set to 1.6, 0.8, 0.4, and 0.2 for four layers. Initial values of learnable parameters $k_{t}$ and $k_{r}$ are set as 0.0 and -2.5 respectively. More experiment details are presented in Appendix.

\textbf{Evaluation metrics.} We follow protocols of DGR \cite{choy2020deep} to evaluate our RegFormer with three metrics: (1) Relative Translation Error (RTE). (2) Relative Rotation Error (RRE). (3) Registration Recall (RR). RR is defined when RRE and RTE are within a certain threshold.


\setlength{\tabcolsep}{0.8mm}
\begin{table}[t]
	\centering
	\footnotesize
	\begin{center}
		\resizebox{1.0\columnwidth}{!}
		{
			\begin{tabular}{l|cc|cc|c|c|c}
				\toprule
				&    \multicolumn{2}{c|}{RTE(m)} &\multicolumn{2}{c|}{RRE(\degree)} &  & & \\ 
				
				
				\multirow{-2}{*}{\begin{tabular}[c]{@{}c@{}}Method \end{tabular}}
				&  AVG  & STD   & AVG    & STD        &\multirow{-2}{*}{\begin{tabular}[c]{@{}c@{}}RR(\%) \end{tabular}}   & \multirow{-2}{*}{\begin{tabular}[c]{@{}c@{}}Time(ms)/Points \end{tabular}} &\multirow{-2}{*}{\begin{tabular}[c]{@{}c@{}}NT(ms) \end{tabular}}\\
				\hline\hline
				\noalign{\smallskip}

                FGR \cite{zhou2016fast}
                &0.93 &0.59
                &0.96 &0.81
                &39.4\%&506.1/11445&44.22\\

                RANSAC \cite{fischler1981random}
                &0.13 &0.07
                &0.54 &0.40
                &91.9\% &549.6/16384&33.54\\

                \cline{1-8}\noalign{\smallskip}

                DCP \cite{wang2019deep}    
				&1.03 & 0.51 
				&2.07 & 1.19 
				&47.3\% & 46.4/1024&45.31\\ 

                IDAM \cite{li2020iterative}    
				&0.66 & 0.48 
				&1.06 & 0.94 
				&70.9\% & 33.4/4096&8.15\\ 

                FMR \cite{huang2020feature}    
				&0.66 & 0.42 
				&1.49 & 0.85 
				&90.6\% & 85.5/12000 &7.13\\ 

                DGR \cite{choy2020deep}    
				&0.32 & 0.32 
				&0.37 & 0.30 
				&98.7\% & 1496.6/16384&91.35\\ 
	
				HRegNet \cite{lu2021hregnet}      
				&0.12 & 0.13
				&0.29 & 0.25
				&99.7\%    & 106.2/16384&6.48\\


                \rowcolor{gray!30}Ours
                &\bf0.08 &\bf0.11
                &\bf 0.23 &\bf0.21
                &\bf{99.8\%} &\bf{98.3/120000}&\bf{0.82}\\
                \bottomrule
			\end{tabular}
		}
	\end{center}
    \vspace{-8pt}
	\caption{Comparison with state-of-the-art. The best performance is highlighted in bold. Registration Recall (RR) is defined as the success ratio where RRE$<$ 5\degree and RTE$<$ 2m. `NT' means normalized time per thousand points.}
	\label{table:2m}

\end{table}

\begin{figure}[t]
  \begin{center}
		\resizebox{1.0\columnwidth}{!}
		{
			\includegraphics[scale=1.00]{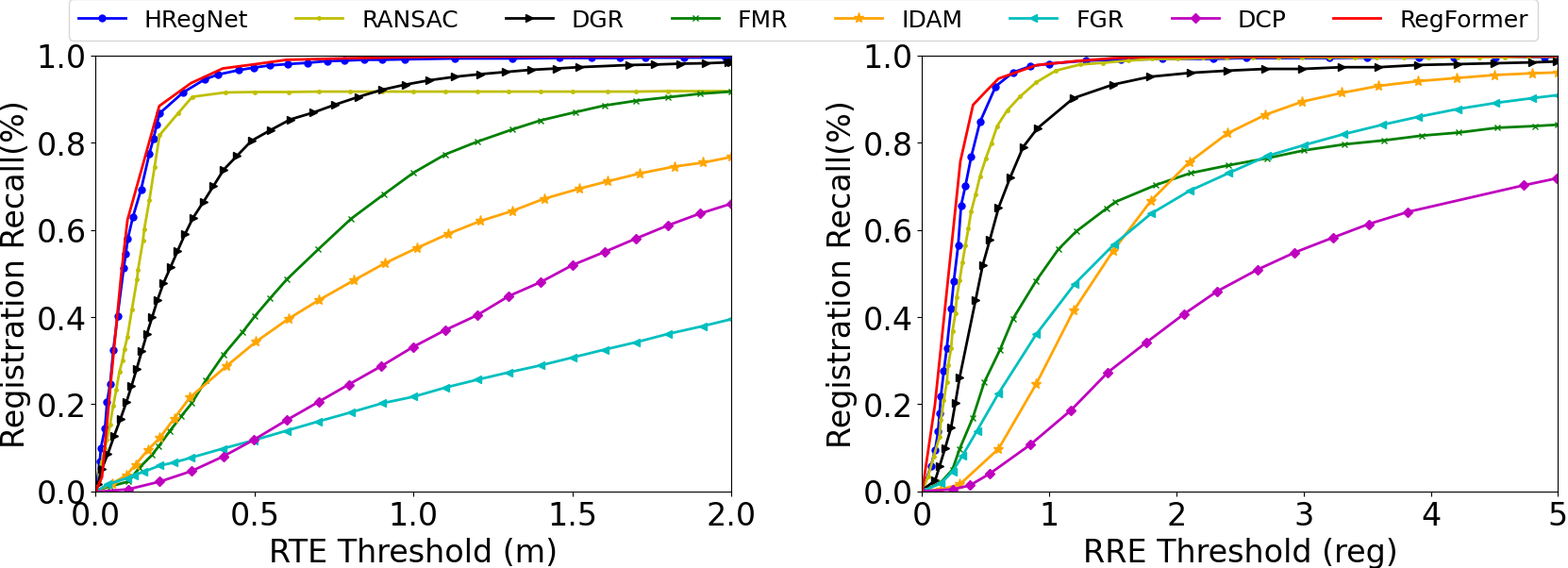}}
	\end{center}
  \vspace{-14pt}
  \caption{Registration recall with different RRE and RTE thresholds on the KITTI dataset.}
  \vspace{-8pt}
  \label{fig:rr}
\end{figure}

\subsection{KITTI Benchmark}
KITTI odometry dataset is composed of 11 sequences (00-10) with ground truth poses. Following the settings in \cite{choy2019fully,choy2020deep}, we use 00-05 for training, 06-07 for validation, and 08-10 for testing. Also, the ground truth poses of KITTI are refined with ICP \cite{choy2019fully,bai2020d3feat}. 

\textbf{Comparison with state-of-the-art.} Following \cite{lu2021hregnet}, the current frame and the 10th frame after it are used to form input point pairs. We choose both traditional and learning-based registration methods for comparison. For classical methods, our network is superior to FGR \cite{zhou2016fast} by a large margin in both accuracy and efficiency. Compared with RANSAC \cite{fischler1981random}, our RegFormer has a 7.9\% higher RR. Also, RANSAC suffers from much lower efficiency (five times more total time than ours) due to the slow convergence. With respect to learning-based methods, RegFormer is compared with a series of state-of-the-art. DCP \cite{wang2019deep}, IDAM \cite{li2020iterative}, and FMR \cite{huang2020feature} are all feature-based registration networks that extract local descriptors. DGR \cite{choy2020deep} achieves competitive performance in indoor scenes with global features. As in Table~\ref{table:2m}, our RegFormer has much lower RRE and RTE, higher RR than all the above learning-based methods without designing discriminative descriptors. HRegNet \cite{lu2021hregnet} is recent CNN-based SOTA for outdoor large-scale scenes. Our RegFormer is more accurate in terms of all metrics and has a 7.2\% efficiency improvement than theirs. As for efficiency, both total time and normalized time are given. Normalized time is calculated by the processing speed per thousand points. As illustrated in Table~\ref{table:2m}, our RegFormer can process large-scale points with the highest average efficiency (0.82ms). Registration recalls with different RRE and RTE thresholds are also displayed in Fig.~\ref{fig:rr}, which proves that our RegFormer is extremely robust to various threshold settings.


\textbf{Comparison with RANSAC-based models.} RANSAC is a commonly employed estimator for filtering outliers. With no need for RANSAC, our RegFormer leverages the attention mechanism for improving the resilience to outliers by learning global features. Its effectiveness is demonstrated by the comparison with RANSAC-based methods in Table~\ref{table:ransac}. We follow settings in \cite{qin2022geometric} using input point pairs at least 10m away and setting the RTE threshold as 2m. Also, all methods are divided into two categories in terms of different backbones: CNN and Transformer. Our network is on par with all SOTA CNN-based works including 3DFeatNet \cite{yew20183dfeat}, FCGF \cite{choy2019fully}, D3Feat \cite{bai2020d3feat}, CoFiNet \cite{yu2021cofinet}, Predator \cite{huang2021predator}, and SpinNet \cite{ao2021spinnet}. Notably, although Predator has 1.6 cm lower RTE than ours due to well-designed local descriptors, it has $11.1\%$ larger RRE and 52$\times$ more runtime compared with ours. In terms of transformer-based networks, GeoTransformer \cite{qin2022geometric} introduces geometric features into transformer and has a marginally smaller RTE, but it has a higher RRE and obvious efficiency decline. Our efficient RegFormer has a $16\times$  speed-up compared with theirs.
\setlength{\tabcolsep}{1.3mm}
\begin{table}[t]
	\centering
	\footnotesize
	\begin{center}
		\resizebox{1.0\columnwidth}{!}
		{
			\begin{tabular}{l|c|c|c|c|c}
				\toprule


				Method&Backbone &RTE(cm) &RRE(\degree) &RR(\%) &Time(s) \\
				\hline\hline
				\noalign{\smallskip}

                3DFeat-Net \cite{yew20183dfeat}&CNN
                &25.9 &0.25
                &96.0\%&3.4\\
          
				FCGF \cite{choy2019fully}  &CNN 
				&9.5 & 0.30 
				&96.6\% &3.4 \\ 

                D3Feat \cite{bai2020d3feat} &CNN
				&7.2 & 0.30
				&\bf{99.8\%} &3.1\\ 
                
                Predator \cite{huang2021predator} &CNN
				&6.8 & 0.27
				&\bf{99.8\%} &5.2\\
                    
				CoFiNet \cite{yu2021cofinet}   &CNN 
				&8.5  &0.41
				&\bf{99.8\%} &1.9\\

                SpinNet \cite{ao2021spinnet}   &CNN
				&9.9  &0.47
				&99.1\% &60.6\\

                
				GeoTransformer \cite{qin2022geometric}   &Transformer
				&7.4  &0.27
				&\bf{99.8\%} &1.6\\


                \rowcolor{gray!30}Ours &Transformer
                &8.4 &\bf0.24
                &\bf{99.8\%}&\bf0.1\\
              
                \bottomrule
			\end{tabular}
		}
	\end{center}
    \vspace{-8pt}
	\caption{Comparison with RANSAC-based networks. The best performance is highlighted in bold. RR is defined as the success ratio where RRE$<$ 5\degree and RTE$<$ 2m.}
	\vspace{-8pt}		
	\label{table:ransac}
\end{table}

\begin{figure*}
 \centering
 \includegraphics[width=1.0\linewidth]{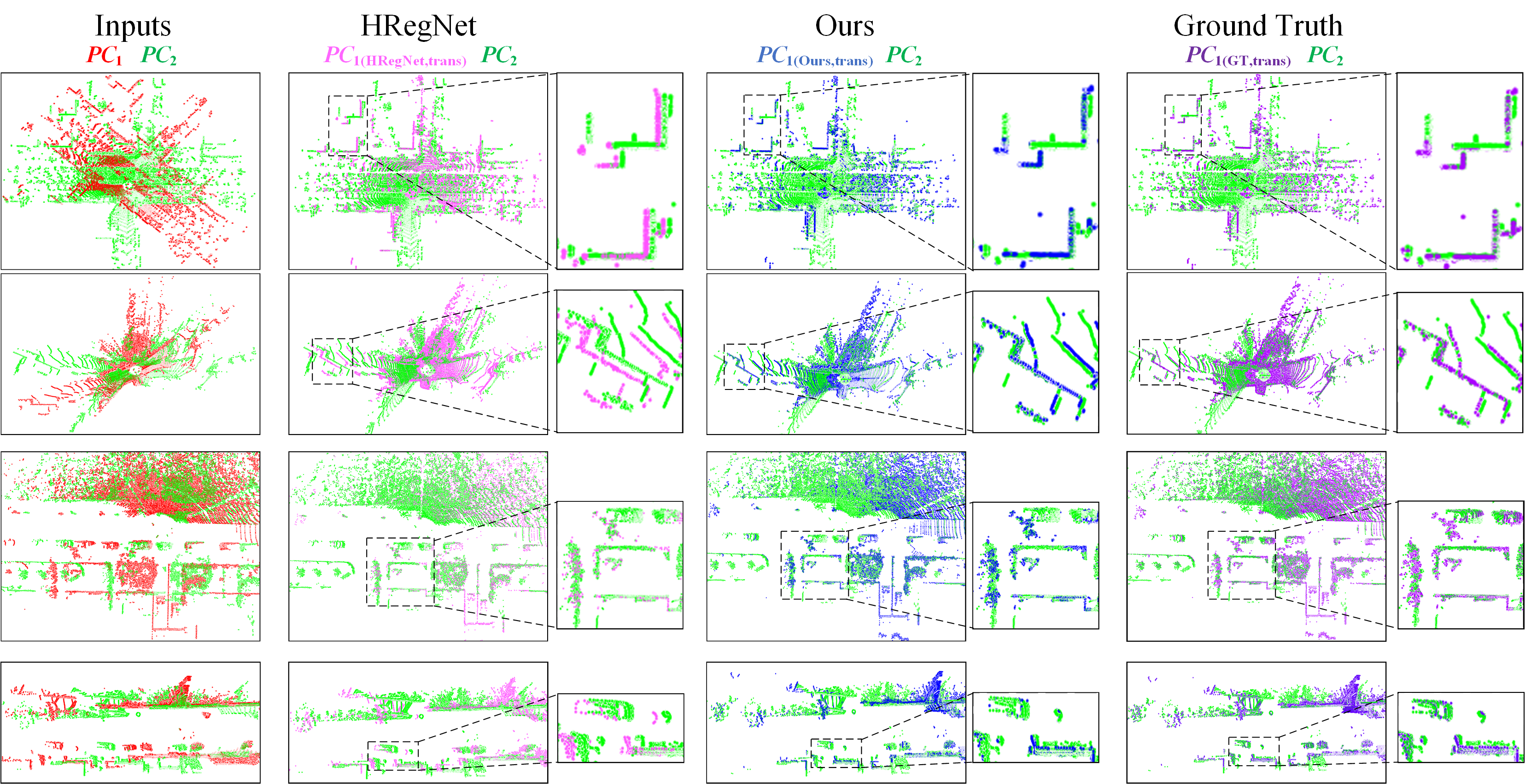}
 \vspace{-14pt}
 \caption{Low-overlap registration results. Point clouds colored red and green indicate the input source and target point clouds. Transformed source points by the estimated pose of ours and HRegNet are colored blue and pink respectively. The ground truth is colored purple. Our RegFormer can align low-overlap input points even with large translations (upper two rows) or rotations (lower two rows).}
 \vspace{-8pt}
 \label{fig:overlap}
\end{figure*}
\vspace{-14pt}
\subsection{NuScenes Benchmark}
We further evaluate our RegFormer on NuScenes. It consists of 1000 scenes, including 850 scenes for training and validation, and 150 scenes for testing. Following \cite{lu2021hregnet}, we use 700 scenes for training, 150 scenes for validation, and 10th frame away point pairs. 

\textbf{Comparison with state-of-the-art.} As illustrated in Table~\ref{table:nuscenes}, our registration accuracy outperforms all classical works and most learning-based ones. Our RegFormer has a 39\% RR improvement compared with RANSAC with only 30\% of their runtime. For learning-based methods, our model is superior to DCP \cite{wang2019deep}, IDAM \cite{li2020iterative}, and FMR \cite{huang2020feature} by a large margin. Compared with HRegNet \cite{lu2021hregnet}, our RegFormer has the same 99.9\% recall. Because the feature extraction section is finely pre-trained, it has 0.02m lower RTE than ours, but more than two times RRE (0.45\degree) instead. Moreover, their efficiency is lower than ours.

\setlength{\tabcolsep}{0.8mm}
\begin{table}[t]
	\centering
	\footnotesize
	\begin{center}
		\resizebox{1.0\columnwidth}{!}
		{
			\begin{tabular}{l|c|c|c|c|c}
				\toprule

				
				Method &RTE(m) &RRE(\degree) &Recall(\%) &Time/Points&NT(ms) \\
				\hline\hline
				\noalign{\smallskip}

                FGR \cite{zhou2016fast}
                &0.71 
                &1.01 
                &32.2\%&284.6/11445&24.87\\

                RANSAC \cite{fischler1981random}
                &0.21 
                &0.74 
                &60.9\% &268.2/8192&32.74\\

                \cline{1-6}\noalign{\smallskip}

                DCP \cite{wang2019deep}    
				&1.09  
				&2.07  
				&58.6\%  &45.5/1024&44.43 \\ 

                IDAM \cite{li2020iterative}    
				&0.47  
				&0.79 
				&88.0\% &32.6/4096&7.96 \\ 

                FMR \cite{huang2020feature}    
				&0.60  
				&1.61  
				&92.1\% &61.1/12000&5.09 \\

                DGR \cite{choy2020deep}    
				&0.21 & 0.48  
				&98.4\% &523.0/8192&63.84\\ 

                HRegNet \cite{lu2021hregnet}      
				&\bf0.18 
				&0.45
				&\bf99.9\%   &87.3/8192&10.66\\

                \rowcolor{gray!30}Ours
                &0.20
                &\bf{0.22} 
                &\bf{99.9\%} &\bf{85.6/50000}&\bf{1.71}\\
                \bottomrule
			\end{tabular}
		}
	\end{center}
    \vspace{-8pt}
	\caption{Quantitative results on NuScenes. The best performance is in bold. `NT' means normalized time per thousand points.}
	\label{table:nuscenes}
\end{table}

\begin{figure}
  \centering
  \includegraphics[width=1.00\linewidth]{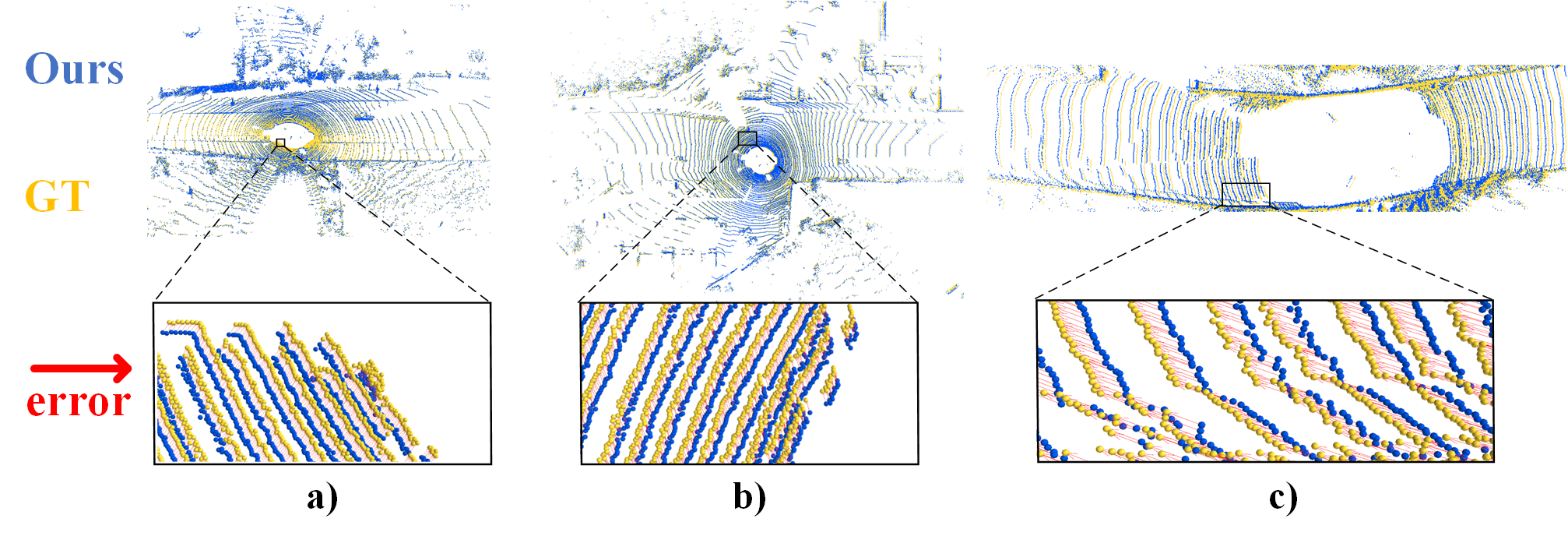}
  \vspace{-14pt}
  \caption{Visualization of registration errors. Point clouds colored yellow and blue indicate transformed source points by the ground truth (GT) and our estimated pose. Registration errors are visualized by a red vector pointing from estimated points to the GT.}
  \vspace{-8pt}
  \label{fig:error}
  
\end{figure}


\subsection{Qualitative Visualization}
\textbf{Low-overlap Registration.} Fig.~\ref{fig:overlap} selects four challenging samples of registration results on the KITTI dataset. Our RegFormer can effectively align source and target point clouds even though they originally have large translations or rotations with low overlap. Also, our RegFormer has higher registration accuracy compared with HRegNet \cite{lu2021hregnet}. 

\textbf{Visualization of the registration errors.} To further study the source of errors, error vectors are visualized in Fig.~\ref{fig:error}. Interestingly, our registration errors are also influenced by the surroundings due to the data-driven characteristics. When features on both sides of the vehicle are sufficient as Fig.~\ref{fig:error} a) and b), errors are relatively distributed evenly. It can be attributed to the structured buildings around, which offer solid positioning information. However, if there are scarce reference objects besides the car or surrounding features are monotonous as in Fig.~\ref{fig:error} c), errors mainly come from the front and rear directions. 

\subsection{Ablation Study}
In this section, extensive ablation studies are conducted for each designed element. 

\textbf{Hierarchical architecture.} We separately output estimated transformation parameters from coarser layers and re-evaluate the metrics. As displayed in Table~\ref{table:ablation1}, rotation and translation are generated from layer 3 (a), layer 2 (b), and layer 1 (c). It is obvious that registration errors get smaller as the transformation is iteratively refined.

\textbf{Projection masks.} The projection mask in our transformer is removed to evaluate its effectiveness. As in Table~\ref{table:ablation2} (a), the whole registration accuracy drops dramatically since numerous invalid pixels are also taken into account.

\textbf{Global modeling capability.} Different from most previous works, our RegFormer focuses on more global features with transformer. The global modeling ability enables our network to sufficiently capture dynamics and recover the occluded objects. To verify this, we replace the feature extraction transformer with CNNs, keeping other components unchanged. As indicated in Table~\ref{table:ablation2} (b), 2D CNN is used to extract local features with the same downsampling scale. PointNet++ \cite{qi2017pointnet++} is also utilized to replace our transformer in Table~\ref{table:ablation2} (c). The results show both local features extracted by 2D CNN and Pointnet++ have larger registration errors since their receptive fields are constrained. 

\textbf{Bijective Association Transformer (BAT).} In this paper, cross-attention is leveraged to exchange information between frames in advance. Here, we remove the cross-attention in BAT to quantitatively test the effectiveness. Table~\ref{table:ablation2} (d) shows that registration errors become double without cross-attention module. Also, the all-to-all point grouping strategy is extremely crucial to find reliable points and reduce mismatches as in Table~\ref{table:ablation2} (e). Furthermore, we conduct experiments by replacing BAT with the cost volume mechanism \cite{wang2021pwclo, wang2022matters}, which is commonly used for consecutive frame association. From the results in Table~\ref{table:ablation2} (f) (g), we can witness at least 0.1m larger RTE and 6.4\% RR drop.

\setlength{\tabcolsep}{1.3mm}
\begin{table}[t]
	\centering
	\footnotesize
	\begin{center}
		\resizebox{1.0\columnwidth}{!}
		{
			\begin{tabular}{l|c|c|c}
				\toprule


				Model &RTE(m) &RRE(\degree) &Recall(\%) \\
				\hline\hline
				\noalign{\smallskip}

                (a) Transformation from layer 3 
                &0.96 $\pm$ 0.42& 1.49 $\pm$ 1.03
                &64.2\%\\
          
				(b) Transformation from layer 2 
				&0.75 $\pm$ 0.44 &1.31 $\pm$0.85
				&79.9\%  \\ 

                (c) Transformation from layer 1 
				 &0.38 $\pm$ 0.30& 0.88 $\pm$ 0.82
                &96.8\%\\

                \cline{1-4}\noalign{\smallskip}
                Ours (from layer 0)
                &\bf{0.08 $\pm$ 0.11} &\bf{0.23 $\pm$ 0.21}
                &\bf{99.8\%}\\

                \bottomrule
			\end{tabular}
		}
	\end{center}
    \vspace{-8pt}
	\caption{Ablation studies of the hierarchical architecture. }
	\label{table:ablation1}
\end{table}
\setlength{\tabcolsep}{0.8mm}
\begin{table}[t]
	\centering
	\footnotesize
	\begin{center}
		\resizebox{1.0\columnwidth}{!}
		{
			\begin{tabular}{l|c|c|c}
				\toprule


				Model &RTE(m) &RRE(\degree) &Recall(\%) \\
				\hline\hline
				\noalign{\smallskip}

          



                (a) w/o projection mask
                &0.22 $\pm$ 0.18&0.52 $\pm$ 0.53
                &98.1\%\\

                \cline{1-4}\noalign{\smallskip}

                (b) replace transformer with 2D CNN
                &0.57 $\pm$ 0.42&0.89 $\pm$ 0.81
                &82.2\%\\

                (c) replace transformer with PointNet++ \cite{qi2017pointnet++}
                &0.24 $\pm$ 0.25&0.57 $\pm$ 0.62
                &92.4\%\\

                \cline{1-4}\noalign{\smallskip}

                (d) w/o cross attention in BAT
                &0.19$\pm$ 0.17&0.48 $\pm$ 0.44
                &98.7\%\\

                (e) w/o all-to-all points gathering in BAT
                &0.88$\pm$ 0.46&1.88 $\pm$ 1.02
                &63.3\%\\

                (f) replace BAT with cost volume in \cite{wang2021pwclo}
                &0.75 $\pm$ 0.49&1.20 $\pm$ 0.90 &60.3\%\\

                (g) replace BAT with cost volume in \cite{wang2022matters}
                &0.18 $\pm$ 0.22&0.33 $\pm$ 0.46 &93.4\%\\

                \cline{1-4}\noalign{\smallskip}

                Ours (Full)
                &\bf{0.08 $\pm$ 0.11} &\bf{0.23 $\pm$ 0.21}
                &\bf{99.8\%}\\

                \bottomrule
			\end{tabular}
		}
	\end{center}
    \vspace{-8pt}
	\caption{Ablation studies of components in transformer module. }
	\vspace{-8pt}		
	\label{table:ablation2}
\end{table}

\section{Discussion}
Here, we discuss why our RegFormer has such excellent accuracy. The competitive performance can be basically attributed to the outlier elimination capability and mismatch rejection strategy. We will elaborate on these two mechanisms in detail respectively.

\begin{figure}
  \centering
  \includegraphics[width=1.00\linewidth]{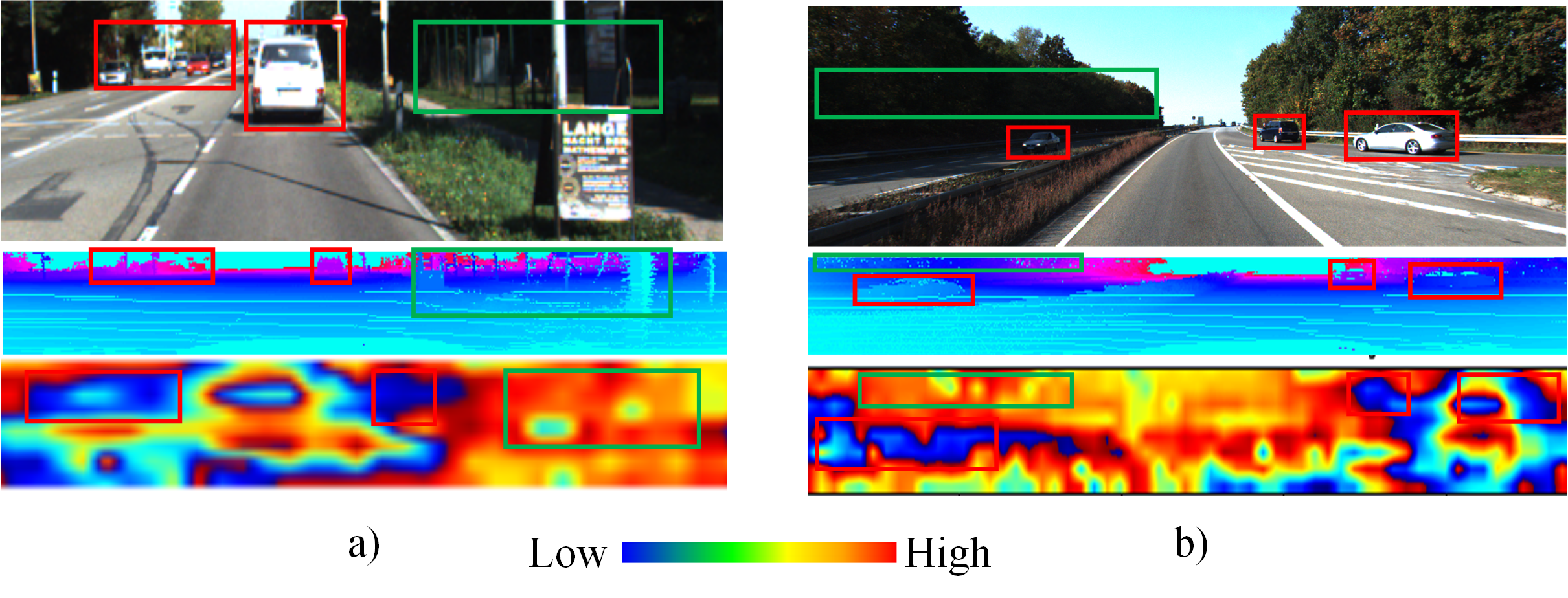}
  \vspace{-14pt}
  \caption{Visualization of attention weights. We give two samples here, where the first two rows respectively represent corresponding pictures and projected point clouds. Attention weights are visualized in the last row. Dynamic objects (red box) have lower attention weights, and static objects (green box) have higher weights.}
  \label{fig:weight}
\end{figure}

\textbf{Global modeling ability to filter outliers.} Transformer can learn patch similarity globally while dynamics and occlusion have inconsistent global motion. So, our transformer-based pipeline can effectively recognize and eliminate interference from these objects by paying less attention to these patches as in Fig.~\ref{fig:weight}. In this case, our RegFormer can maintain high registration accuracy without robust estimators like RANSAC.

\begin{figure}
  \centering
  \includegraphics[width=1.00\linewidth]{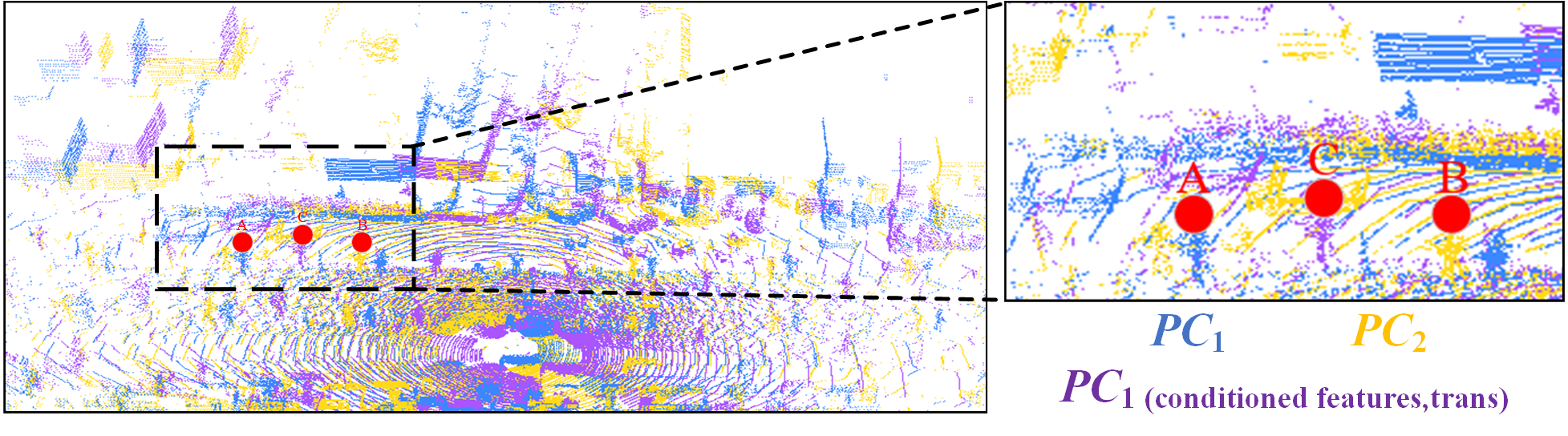}
  \vspace{-14pt}
  \caption{Visualization of the cross attention mechanism in BAT. Points A and B are corresponding points respectively in the source and target frame. Point C is the transformed position of A by conditioned features after cross-attention block.}
  \vspace{-8pt}
  \label{fig:visual2}
\end{figure}

\textbf{Cross-attention mechanism for reducing mismatches.} In our Bijective Association Transformer module, a cross-attention block is first applied to exchange information and embed motion between two frames. Here, we remove the rest parts of BAT, leveraging only conditioned features from the cross-attention block to generate a directionally correct but not precise transformation. Then, it is used to transform input point clouds as in Fig.~\ref{fig:visual2} (purple). For each point in the source point cloud (blue), its corresponding point in the target one (yellow) is originally almost 10m away. Cross-attention can effectively shorten this distance between two frames by learning preliminary motion embeddings. 


\section{Conclusion}
In this paper, we proposed a transformer-based large-scale registration network. Global features are extracted by transformer to filter outliers. To cope with the irregularity and sparsity of raw point clouds, we leverage cylindrical projection to organize them orderly and present a projection mask to remove invalid pixels. Furthermore, a bijective association transformer, including cross-attention-based preliminary information exchange and all-to-all point gathering, is designed for reducing mismatches. The whole model is RANSAC-free, high-accuracy, and extremely efficient. 

{\bf \small  Acknowledgement.} {\small This work was supported in part by the Natural Science Foundation of China under Grant 62225309, 62073222, U21A20480, and U1913204. Authors gratefully appreciate the contribution of Qirong Liu from CUMT and Yu Zheng from SJTU. }

{{\small
\bibliographystyle{ieee_fullname}
\bibliography{iccv23}
}

\end{document}